\documentclass[10pt,twocolumn,letterpaper]{article}

\usepackage[pagenumbers]{cvpr} % To force page numbers, e.g. for an arXiv version

\definecolor{cvprblue}{rgb}{0.21,0.49,0.74}
\usepackage[pagebackref,breaklinks,colorlinks,allcolors=cvprblue]{hyperref}

\usepackage{multirow}
\usepackage{multicol}
\usepackage{nicefrac}
\usepackage{makecell}

\def\paperID{25114} % *** Enter the Paper ID here
\def\confName{CVPR}
\def\confYear{2026}

\title{LaverNet: Lightweight All-in-one Video Restoration \\via Selective Propagation}

\author{Haiyu Zhao, Yiwen Shan, Yuanbiao Gou, Xi Peng\\
College of Computer Science, Sichuan University.\\
\tt\small\{haiyuzhao.gm, yiwenshan, gouyuanbiao, pengx.gm\}@gmail.com
}

\begin{document}
\maketitle

\begin{abstract}
% Video restoration aims to recover high-quality videos from degraded inputs affected by noise, blur, and compression. Recent studies have explored all-in-one video restoration, which handles multiple degradations with a unified model. However, these approaches still face two challenges when dealing with time-varying degradations. First, the degradation can dominate temporal modeling, causing the model to focus on artifacts rather than the video content. Second, current methods typically rely on large models to handle all-in-one restoration, resulting in high computational cost. To address these challenges, we propose a lightweight all-in-one video restoration network, LaverNet, with only 362K parameters. To mitigate the impact of degradations on temporal modeling, we introduce a novel propagation mechanism to dynamically selects and transmits only the degradation-agnostic features across frames. Despite using less than 1\% of the parameters of previous all-in-one video restoration method ViWS-Net, LaverNet achieves comparable or even superior performance across all benchmarks, demonstrating both efficiency and effectiveness.
Recent studies have explored all-in-one video restoration, which handles multiple degradations with a unified model. However, these approaches still face two challenges when dealing with time-varying degradations. First, the degradation can dominate temporal modeling, confusing the model to focus on artifacts rather than the video content. Second, current methods typically rely on large models to handle all-in-one restoration, concealing those underlying difficulties. To address these challenges, we propose a lightweight all-in-one video restoration network, LaverNet, with only 362K parameters. To mitigate the impact of degradations on temporal modeling, we introduce a novel propagation mechanism that selectively transmits only degradation-agnostic features across frames. Through LaverNet, we demonstrate that strong all-in-one restoration can be achieved with a compact network. Despite its small size, less than 1\% of the parameters of existing models, LaverNet achieves comparable, even superior performance across benchmarks.
\end{abstract}

%%%%%%%%% BODY TEXT
\section{Introduction}
\label{sec:intro}

% Video restoration aims to reconstruct high-quality videos from low-quality inputs degraded by factors such as noise, blur and compression. In contrast to image restoration that relies solely on the spatial information within a single image, video restoration leverages the complementary information across frames. By exploiting this temporal redundancy, existing video restoration methods have achieved remarkable success, showing clear advantages in both performance and efficiency over image restoration approaches. However, most video restoration methods are designed for specific types of degradation and struggle to handle the real-world scenarios, where videos are usually corrupted by multiple degradations.
Video restoration aims to reconstruct high-quality videos from low-quality inputs degraded by factors such as noise, blur and compression. Unlike image restoration methods~\cite{Image1, Image2, Image3, Image4, Image5, Image6}, which relies solely on spatial information, video restoration methods~\cite{Zuo1Deblur, Zuo3Denoising, Zuo4Desmoking, Zhang2Restoration, Zhang3Denoising, Ren1Deblurring} exploits temporal redundancy across frames, achieving superior performance and efficiency. However, most existing methods are tailored to specific degradations and struggle with real-world scenarios involving multiple degradations.

% To address this issue, several studies have explored all-in-one video restoration, which aims to handle multiple degradations using a unified model. Some methods can remove different degradations across multiple videos, but they generally assume that each video contains only a single type of degradation. In contrast, AverNet tackles a more challenging scenario where degradations vary over time within a single video, enabling the model to recover videos from dynamic degradations. Thanks to these efforts, all-in-one video restoration has been evolving toward more realistic and practical directions. 
\begin{figure}[htbp]
\includegraphics[width=\linewidth]{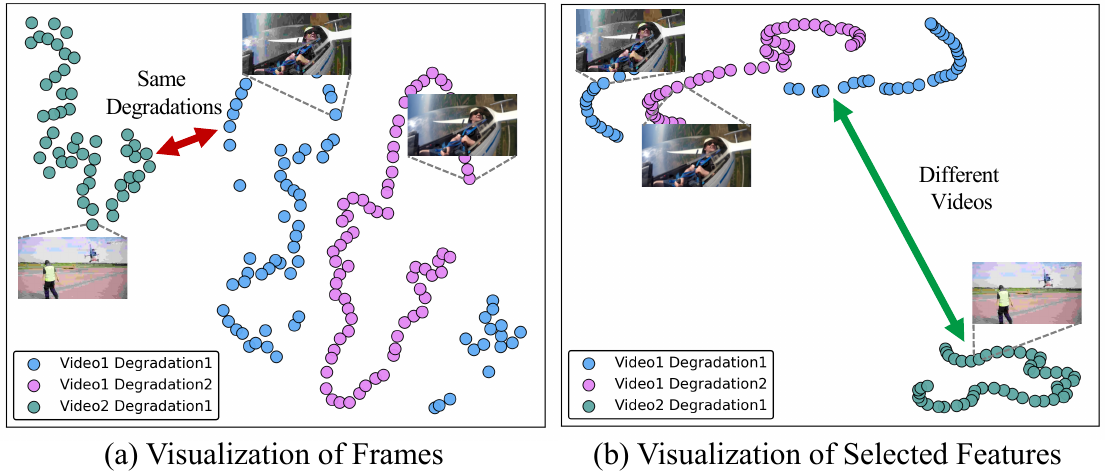}
\caption{Visualization of (a) the frame features and (b) the selected features using our proposed Selective Propagation Mechanism (SPM). It can be observed that SPM selectively propagates degradation-agnostic features, facilitating more efficient propagation of temporally correlated features.
% Blue and purple points represent the same video under different degradations, while blue and green points represent differnt videos with the same degradation. 
}
\label{Fig1}
\end{figure}

To address this issue, recent studies have explored all-in-one video restoration~\cite{ViWS-Net,CUDN,Diff-TTA,AverNet,AVR1,AVR2}, which aims to handle multiple degradations within a single unified model. While these approaches have achieved promising results, they generally assume that each video is affected by only one type of degradation. In contrast, the more recent AverNet~\cite{AverNet} tackles a more challenging scenario of temporally varying degradations within a single video, advancing toward a more realistic and practical all-in-one setting.

% However, these methods still suffer from two major limitations. On the one hand, the degradations could dominate temporal modeling, causing the model to focus on degradations rather than the actual video content. As shown in Fig. 1 (left), videos with the same type of degradation appear even more similar to each other than those sharing the same content, suggesting that degradations interfere with temporal modeling and hinder the model from learning video content. On the other hand, these methods contain a large number of parameters and incur long runtime, make them impractical for real-world deployment, especially in resource-constrained environments.
Despite these advances, existing methods still face two major challenges when dealing with time-varying degradations. First, degradations such as noise, blur, or compression can dominate temporal modeling, causing the model to focus on artifacts rather than the underlying video content. As shown in Fig.~\ref{Fig1}(a), videos affected by the same type of degradation often appear similarly alike, sometimes even more so than videos depicting the same scene, illustrating how degradations can confuse the model and hinder temporal feature propagation. Second, most methods address the all-in-one challenges by scaling up model size, concealing the underlying difficulties rather than providing more insightful or adaptive solutions.

% To address these limitations, we propose a lightweight all-in-one video restoration network. To achieve efficiency, we design a compact architecture and efficient modules with only about 362K parameters, accounting for merely 0.8\% of the all-in-one video restoration method ViWS-Net~\cite{ViWS-Net}, and 1\% of the all-in-one image restoration method PromptIR~\cite{PromptIR}. For temporal modeling, we introduce a propagation module inspired by the selective state update in Mamba. Our module dynamically selects and propagates informative degradation-agnostic features, thereby enabling robust temporal modeling. As illustrated in Fig. 1 (right), the selected features of the same video under different degradations become significantly closer to each other, indicating the model effectively captures the degradation-agnostic video content.
In this work, we would demonstrate that, with deep insights and clever designs, a small and lightweight model can achieve remarkable performance. To this end, we introduce a novel propagation mechanism inspired by the selectivity principle in Mamba~\cite{Mamba}. Unlike conventional methods~\cite{BasicVSR, BasicVSR++, RVRT} that propagate features indiscriminately, the new mechanism dynamically selects and transmits only the most informative, degradation-agnostic features across frames. This targeted propagation alleviates confusion caused by varying degradations and enables robust temporal modeling with a lightweight architecture. As shown in Fig.~\ref{Fig1}(b), features from the same video under different degradations are drawn significantly closer in feature space, demonstrating that our mechanism selectively captures the underlying video content while ignoring corrupting artifacts.

% Building upon the above design, we construct LaverNet with two key components, \ie, the Selective Propagation Module (SPM) and the Lightweight Enhancement Module (LEM), which are responsible for propagating temporal information and adaptively handling various degradations, respectively. Specifically, SPM estimates a selection matrix from the current frame to identify degradation-agnostic features, which are then integrated into the intermediate temporal representation and propagated to the next frame to facilitate subsequent restoration. LEM enhances each frame through a combination of two dense blocks and a degradation-guided attention block. The dense block captures fine-grained spatial details, while the attention block extracts degradation cues from the input frame and leverages them to adaptively handle the degradations.
Building on this propagation mechanism, we propose the Lightweight All-in-one VidEo Restoration Network (LaverNet), which employs a compact architecture and carefully designed modules. The entire network contains only about 362K parameters, just 0.8\% of the all-in-one video restoration network ViWS-Net~\cite{ViWS-Net}. To achieve this, we introduce two key components: the Lightweight Enhancement Module (LEM) and the Selective Propagation Module (SPM). LEM efficiently handles each type of degradation in every frame, while SPM selectively propagates degradation-agnostic temporal information to next frame.

Specifically, LEM focuses on per-frame enhancement by combining two dense blocks and a degradation-guided attention block. The dense blocks capture fine-grained spatial details, while the attention block extracts degradation cues from each frame and adaptively compensates for them, allowing LEM to efficiently handle various types of degradation. In contrast, SPM selectively propagates degradation-agnostic temporal information by estimating a selection matrix from the current frame, which identifies informative features to be integrated into the temporal representation and propagated to the next frame for subsequent restoration.

To summarize, the contributions and novelty of this work are as below:
\begin{itemize}
\item We propose LaverNet, a lightweight all-in-one video restoration model that achieves strong restoration performance with only 362K parameters, significantly smaller than existing all-in-one models.
\item We design SPM, inspired by the selectivity principle in Mamba, to mitigate degradation interference by selectively propagating useful temporal cues while forgetting corruption across frames.
\item Extensive experiments show that LaverNet achieves superior performance across diverse degradations and datasets, while using only about 1\% parameters of existing all-in-one video restoration models.
\end{itemize}

\section{Related Works}

In this section, we present an overview of the related works, including all-in-one image restoration, lightweight video restoration, and all-in-one video restoration.

\subsection{All-in-one Image Restoration}

All-in-one image restoration~\cite{AIR1,AIR2,AIR3,AIR4,AIR5} aims to recover images affected by diverse degradations using a single unified model. Instead of training separate models for different degradations, recent approaches focus on learning degradation-specific representations that could adaptively guide the restoration process.

For example, AirNet~\cite{AirNet} introduces contrastive learning to acquire degradation-aware representations, which are subsequently incorporated into the network for all-in-one image restoration. PromptIR~\cite{PromptIR} introduces learnable prompts to encode degradation-specific priors, enabling adaptive restoration under diverse conditions. ProRes~\cite{ProRes} introduces additional visual prompts to incorporate task-specific information and utilize the prompts to guide the network for restoration. MPerceiver~\cite{MPerceiver} adopts a multimodal prompt-learning approach, utilizing generative priors from Stable Diffusion~\cite{StableDiffusion} to achieve high-fidelity all-in-one restoration. LoRA-IA~\cite{LoRAIR} further employs a routing strategy and mixture-of-experts architecture to dynamically handle multiple degradations. PromptRestorer~\cite{PromptRestorer} explicitly modulates degradation representations, enabling the network to more effectively perceive and exploit degradation priors for restoration.

Despite their success, existing all-in-one image restoration methods typically involve a large number of parameters, often exceeding 10M. In contrast, our method contain only about 362K parameters, which is significantly fewer than these models. For instance, our method requires only approximately 1.5\% of the parameters of PromptRestorer and around 2.8\% of its runtime.

\subsection{Lightweight Video Restoration}

Lightweight video restoration~\cite{FastDVDNet, DVDNet, RFDA, STDF, EvrNet} focuses on efficiently recovering high-quality videos while minimizing model complexity and computational cost. These methods aim to exploit temporal redundancy across frames and enhance spatial details without relying on large networks.

For example, RFDA~\cite{RFDA} employs recursive fusion to capture temporal dependency and introduces deformable spatialtemporal attention to aggregate information from multiple frames. STDF~\cite{STDF} applies spatial-temporal deformable convolutions to compensate for motion and integrate features from neighboring frames. EvrNet~\cite{EvrNet} designs a lightweight network with three stages, \ie, alignment, differential, and fusion, which align frame, learn high-frequency representations, and aggregate temporal information, respectively.

Although these methods are lightweight and effective, they overlook the interference of degradations during temporal modeling. Consequently, their performance decreases significantly when handling time-varying degradations, as observed for RFDA and STDF in our experiments. In contrast, our method explicitly selects and propagates degradation-agnostic features, enabling robust temporal modeling. This design allows our network to achieve strong restoration performance while maintaining high efficiency.

\begin{figure*}[htbp]
\includegraphics[width=\linewidth]{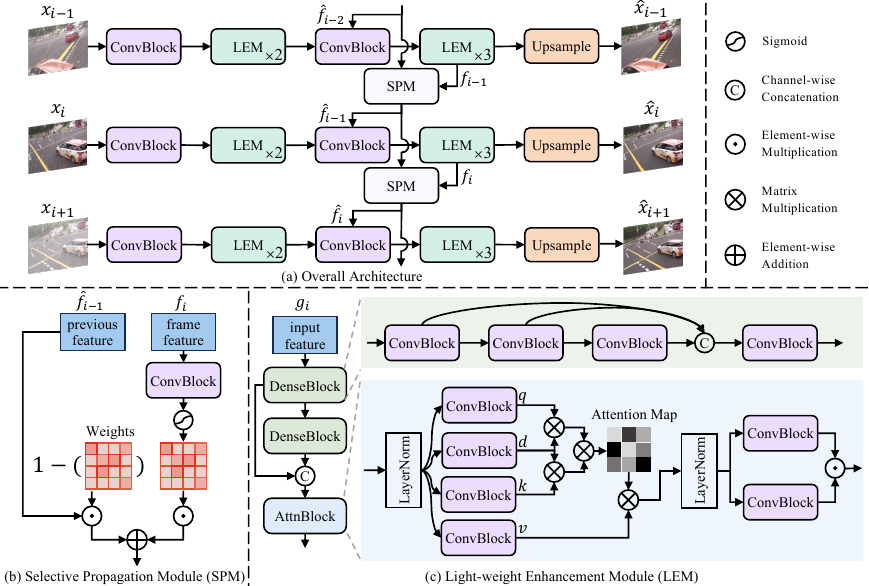}
\caption{Architecture overview. (a) Overall architecture of LaverNet, comprising two key modules: (b) Selective Propagation Module (SPM), which propagates degradation-agnostic temporal information, and (c) Lightweight Enhancement Module (LEM), which efficiently handles each type of degradation in every frame.}
\label{Fig2}
\end{figure*}

\subsection{All-in-one Video Restoration}

Most existing video restoration methods~\cite{Zhang4Denoising, Zuo2Deblur, Zuo5Deblurring, Zhang1Deblurring, Pan1Deblurring, Pan2Deblurring} are designed for a specific type of degradation. In contrast, recent studies~\cite{CUDN,Diff-TTA,ViWS-Net,AverNet} explore all-in-one video restoration, which aims to address multiple degradations with a unified model. For example, CUDN~\cite{CUDN} adaptively estimates degradation representations to guide restoration under various degradation conditions. Diff-TTA~\cite{Diff-TTA} employs test-time adaptation to address multiple degradation by aligning the test data distribution and updating the parameters of pre-trained models accordingly. ViWSNet~\cite{ViWS-Net} introduces degradation messenger tokens to encode degradation-specific information and incorporates them to guide the restoration process. Different from these methods, a recent work, AverNet~\cite{AverNet}, first focuses on the time-varying nature of video degradations and incorporates prompt learning into a recurrent framework to address the time-varying degradations.

Although these all-in-one video restoration methods effectively handle multiple degradations, they often suffer from substantial model complexity and parameter overhead. For instance, both ViWSNet and AverNet require over 40M parameters. In contrast, our method uses only about 362K parameters, which is approximately 0.6\% of ViWSNet. Despite this compact design, it achieves superior restoration performance and requires only one tenth of ViWSNet's runtime, demonstrating both high efficiency and effectiveness.

\section{Proposed Method}

In this section, we elaborate on our proposed LaverNet and its two key modules, namely Selective Propagation Module (SPM) and Lightweight Enhancement Module (LEM). 

\subsection{Overall Architecture}

The overall architecture of LaverNet is shown in Fig.~\ref{Fig2}. For the $i$-th input frame $x_i$, a convolutional block together with two LEMs are first applied to extract shallow spatial features $f_i^1$:
\begin{equation}
    f_i^1 = \text{LEM}_{\times2}(\text{ConvBlock}(x_i)).
\end{equation}
These features are then fused with the propagated features from the previous frame, which are obtained via SPM to selectively propagate contents from the previous frame:
\begin{equation}
    \label{eq:spm}
    \hat{f}_{i-1} = \text{SPM}(f_{i-1}, \hat{f}_{i-2}),
\end{equation}
where $\hat{f}_{i-1}$ is the propagated feature from the previous frame, $f_{i-1}$ is the feature of the previous frame, $\hat{f}_{i-2}$ is the propagated feature from the frame two steps before. Fusion is performed through channel concatenation followed by a convolutional block:
\begin{equation}
    f_i^2 = \text{ConvBlock}(\text{Concat}(f_i^1, \hat{f}_{i-1}))
\end{equation}
where $f_i^2$ is the fused feature, which is subsequently refined by three additional LEMs, \ie,
\begin{equation}
    f_i = \text{LEM}_{\times3}(f_i^2),
\end{equation}
where $f_i$ is the feature of the current frame, later fed into an upsampling block to reconstruct the high-quality output $y_i$, \ie,
\begin{equation}
    y_i = \text{Upsample}(\text{LEM}_{\times3}(f_i)).
\end{equation}
Meanwhile, $f_i$ also serves as the current frame feature to generate the propagated feature $\hat{f}_i$ for the $(i+1)$-th frame, following Eq.~(\ref{eq:spm}). Since $f_i$ is relatively deep and contains minimal degradation, it is well suited for propagating informative, degradation-agnostic contents to subsequent frames.

In the following, we describe the key modules, \ie, SPM and LEM, that enable lightweight, all-in-one video restoration.

\subsection{Selective Propagation Module}

Video degradations pose challenges for consistently capturing the underlying content across frames, as illustrated in Fig.~\ref{Fig1}. To address this issue, we propose the Selective Propagation Module (SPM), which dynamically identifies and propagates degradation-agnostic features across frames, enabling robust temporal modeling.

Specifically, SPM predicts a weight matrix to filter the previous frame’s feature, allowing only relevant temporal information to be passed to the current frame. To be specific, given the feature of the previous frame $f_{i-1}\in\mathbb{R}^{c\times h\times w}$, the weight matrix $w_{i-1}\in\mathbb{R}^{c\times h\times w}$ is computed as:
\begin{equation}
    w_{i-1} = \sigma(\text{ConvBlock}(f_{i-1})),
\end{equation}
where $\sigma$ represents the sigmoid activation function. 

The predicted weight matrix $w_{i-1}$ selectively modulates the previous frame feature $f_{i-1}$, suppressing irrelevant regions while preserving informative contents, and is then combined with the propagated feature from the frame before, $\hat{f}_{i-2}$, as
\begin{equation}
    \hat{f}_{i-1} = (1 - w_{i-1}) \odot \hat{f}_{i-2} + w_{i-1} \odot f_{i-1},
\end{equation}
where $\odot$ denotes element-wise multiplication, and $\hat{f}_{i-2}$ represents the propagated feature from $(i-2)$-th frame.

% \textcolor{blue}{To elaborate on the advantages and innovations of this mechanism, as well as how to enable lightweight, all-in-one video restoration.} 

The design of SPM is inspired by the selectivity principle in Mamba, which dynamically identifies important features for subsequent processing. Motivated by this idea, we for the first time explicitly consider and address the interference of degradations in temporal modeling for video restoration. To this end, we introduce the highly lightweight SPM, which effectively identifies and propagates degradation-agnostic features, as illustrated in Fig.~\ref{Fig1}. Moreover, SPM introduces only 20K additional parameters yet yields substantial performance gains, as shown in Tab.~\ref{Tab:SPM}, demonstrating its effectiveness.

\begin{table}[hbtp]
\caption{Performance w/o and w/ SPM. With only 20.6K parameters, SPM yields notable improvements in PSNR and SSIM.}
\centering
\begin{tabular}{l|r|c|c}
\toprule
 & \multirow{2}{*}{\#Params} & \multicolumn{2}{|c}{DAVIS-test} \\ 
 \cmidrule{3-4} 
 & & PSNR & SSIM \\
 \midrule
 w/o SPM & 342.1K & 32.32 & 0.8896 \\
 w/ SPM & 362.7K & 32.60 & 0.8967\\
 Gains & 20.6K & 0.28 & 0.0071 \\
\bottomrule
\end{tabular}
\label{Tab:SPM}
\end{table}

\subsection{Lightweight Enhancement Module}

To handle different types of degradation in input video, LEM incorporates three lightweight modules: two dense blocks and a attention block. The dense block captures fine-grained details, while the attention block identifies and mitigates degradation patterns. By capturing fine-grained degradation cues, LEM enables the restoration process to be conditioned on the specific degradation present in each frame.

Given the input feature $g_i \in \mathbb{R}^{c \times h \times w}$, the dense block sequentially applies three convolutional layers and integrates their outputs to capture multi-level information. Formally, this process can be expressed as
\begin{equation}
\begin{aligned}
    g^1_i &= \text{ConvBlock}(g_i), \\
    g^2_i &= \text{ConvBlock}(g^1_i), \\
    g^3_i &= \text{ConvBlock}(g^2_i), \\
    g^4_i &= \text{Concat}(g^1_i, g^2_i, g^3_i), \\
    g'_i  &= \text{ConvBlock}(g^4_i),
\end{aligned}
\end{equation}
where $\text{Concat}(\cdot)$ denotes channel-wise concatenation. The output $g'_i$ is then fed into a second dense block, yielding $g''_i$. Finally, the outputs of the two dense blocks are concatenated to form the aggregated feature, \ie,
\begin{equation}
\hat{g}_i = \text{Concat}(g'_i, g''_i).
\end{equation}

The aggregated feature $\hat{g}_i$ is subsequently fed into attention block. In this block, the query $q$, key $k$, value $v$, and degradation cues $d$ are first extracted:
\begin{equation}
    \begin{aligned}
        q &= \text{ConvBlock}_q(\text{LN}(\hat{g}_i)), \\
        k &= \text{ConvBlock}_k(\text{LN}(\hat{g}_i)), \\
        v &= \text{ConvBlock}_v(\text{LN}(\hat{g}_i)), \\
        d &= \text{ConvBlock}_d(\text{LN}(\hat{g}_i)),
    \end{aligned}
\end{equation}
where $\text{LN}(\cdot)$ denotes the layer normalization. The dimensions of $q$, $k$, $v$, and $d$ are $\mathbb{R}^{n\times \frac{c}{n}\times hw}$, where $n$ is the number of attention heads.

The attention block then integrates the degradation cues into the attention calculation to conditionally mitigate the degradations present in each frame:
\begin{equation}
    h_i = \text{Softmax}((q\cdot d^T) \cdot (k\cdot d^T)^T) \cdot v. \\
\end{equation}
To further enhance the representation, $\hat{g}_i$ is passed through a lightweight feed-forward module:
\begin{equation}
    \hat{h}_i = \text{LN}(\text{ConvBlock}_1(h_i))\odot \text{LN}(\text{ConvBlock}_2(h_i)).
\end{equation}

By explicitly incorporating degradation cues into attention computation, the block could dynamically adapt to different degradations. Specifically, the degradation cues are multiplied with the query and key to guide the attention toward degradation-relevant regions, without introducing significant additional parameters or computational cost. As a result, this design enables adaptive handling of various degradations while maintaining high efficiency, achieving lightweight all-in-one video restoration.

\section{Experiment}

In this section, we conduct experiments to assess the proposed method. We first describe the experimental settings, including the datasets, architecture details, and the training details. Then, we present quantitative and qualitative results. Finally, we conduct ablation studies to demonstrate the effectiveness of key modules.

\subsection{Experiment Settings}

\textbf{Architecture Details.} We adopt the same settings across all experiments. Specifically, the number of channels is set to $32$, with 2 LEMs at the beginning of the network and 3 LEMs at the end. To further validate the effectiveness of our method, we construct two variants of our LaverNet by adjusting the channel width. The tiny version (LaverNet-T) has $16$ channels and the large version (LaverNet-L) has $48$ channels.

\noindent\textbf{Training Details.} The experiments are conducted in PyTorch~\cite{PyTorch} framework with NVIDIA GeForce RTX 3090 GPUs and NVIDIA A800 GPUs. Specifically, our model is trained on two RTX 3090 GPUs. For training, we adopt Charbonnier~\cite{Charbonnier} loss and Adam~\cite{Adam} optimizer with $\beta_1=0.9$ and $\beta_2=0.999$. The initial learning rate is set to $2e^{-4}$ for all datasets and gradually decreased to $1e^{-7}$ through the cosine annealing~\cite{CosineAnnealing} strategy. Each training sample consists of $12$ input frames with a resolution of $256\times 256$. The networks are trained for 600K iterations with a batch size of $2$.

\noindent\textbf{Video Datasets.} We conduct our experiments on two synthetic video datasets DAVIS~\cite{DAVIS} and Set8~\cite{Set8}, as used in~\cite{AverNet}. The DAVIS dataset include 90 sequences for training and 30 sequences for testing, each with a resolution of $854\times 480$. The Set8 dataset contains 8 test sequences at $960\times 540$ resolution. All models are trained on the DAVIS training set and evaluated on both the DAVIS test set and Set8.

\subsection{Comparison Experiments}

In this section, we compare our LaverNet with existing state-of-the-art methods, including all-in-one image restoration, video restoration, and all-in-one video restoration approaches. Specifically, the all-in-one image restoration methods include WDiffusion~\cite{WDiffusion}, TransWeather~\cite{TransWeather}, LoRA-IR~\cite{LoRAIR}, and PromptRestorer~\cite{PromptRestorer}. The video restoration methods are EDVR~\cite{EDVR}, RFDA~\cite{RFDA}, STDF~\cite{STDF}, and EvrNet~\cite{EvrNet}, while the all-in-one video restoration method is ViWS-Net~\cite{ViWS-Net}. The image and video restoration methods were trained on the synthesized TUD-Common~\cite{AverNet} dataset from scratch according to the training settings in their papers. Note that the image restoration methods were trained and tested on each frame of the video sequences. 

\begin{table}[hbtp]
\caption{Comparisons of parameters, FLOPs and runtime.}
\label{Tab:paramCompare}
\centering
\resizebox{\linewidth}{!}{
    \begin{tabular}{c|l|r|r|r}
    \toprule
    Type & Method & \#Param & FLOPs & Runtime \\ \midrule
    \multirow{5}{*}{Image Restoration} & WDiffusion & 80.0M & 73.26T & 3956.50s \\
    & TransWeather & 38.1M & 38.88G & 3.32s \\
    & LoRA-IR & 68.9M & 398.97G & 5.45s\\
    & PromptRestorer & 24.5M & 66.62G & 46.93s \\
     \midrule
    \multirow{4}{*}{Video Restoration} & EDVR & 23.6M & 3.30T & 6.78s \\
    & RFDA & 1.3M & 280.73G & 4.03s \\ 
    & STDF & 365.0K & 150.07G & 0.95s\\
    & EvrNet & 78.7K & 14.15G & 0.91s\\
    \midrule
    \multirow{2}{*}{\makecell{All-in-one \\Video Restoration}} & ViWS-Net & 57.8M & 548.88G & 12.20s \\
    & LaverNet (Ours) & 362.7K & 36.90G & 1.32s \\ \bottomrule
    \end{tabular}
}
\end{table}

\begin{table*}[htbp]
\caption{Quantitative results compared with state-of-the-art methods on test sets with different degradation change intervals, where $t$ denotes the interval at which degradations change. The best outcomes are highlighted in \textbf{bold}.}
\label{Tab:quanti-1}
\begin{center}
    \resizebox{\textwidth}{!}{
    \begin{tabular}{l|r|cc|cc|cc|cc|cc|cc}
    \toprule
        \multicolumn{1}{c|}{\multirow{3}{*}{Method}} & \multicolumn{1}{c|}{\multirow{3}{*}{\#Param}} & \multicolumn{6}{c|}{DAVIS-test} & \multicolumn{6}{c}{Set8} \\ \cmidrule{3-14} 
          & & \multicolumn{2}{c|}{t=6} & \multicolumn{2}{c|}{t=12} & \multicolumn{2}{c|}{t=24} & \multicolumn{2}{c|}{t=6} & \multicolumn{2}{c|}{t=12} & \multicolumn{2}{c}{t=24}   \\ \cmidrule{3-14} 
          & & PSNR & SSIM & PSNR & SSIM & PSNR & SSIM & PSNR & SSIM & PSNR & SSIM & PSNR & SSIM\\
    \midrule
    WDiffusion & 80.0M & 31.74 & 0.8768 & 31.79 & 0.8784 & 31.92 & 0.8809 & \textbf{30.31} & 0.8784 & 30.02 & 0.8716 & \textbf{30.82} & 0.8746\\
    TransWeather & 38.1M & 31.11 & 0.8694 & 31.13 & 0.8699 & 31.26 & 0.8741 & 29.24 & 0.8662 & 28.95 & 0.8565 & 29.15 & 0.8632\\
	LoRA-IR & 68.9M & 27.12 & 0.6696 & 26.75 & 0.6518 & 26.76 & 0.6596 & 24.90 & 0.6427 & 24.43 & 0.6188 & 24.98 & 0.6464\\
	PromptRestorer & 24.5M & 31.38 & 0.8707 & 31.46 & 0.8695 & 31.53 & 0.8737 & 29.87 & 0.8663 & 29.50 & 0.8549 & 30.79 & 0.8733\\
    \midrule
    EDVR & 23.6M & 28.70 & 0.7224 & 28.37 & 0.6991 & 29.07 & 0.7289 & 26.75 & 0.7259 & 26.94 & 0.7382 & 28.71 & 0.7675\\
    RFDA & 1.3M & 20.79 & 0.6343 & 20.78 & 0.6317 & 19.31 & 0.5235 & 19.31 & 0.5235 & 19.34 & 0.5257 & 19.29 & 0.5258\\
    STDF & 365.0K & 21.13 & 0.5768 & 21.08 & 0.5594 & 21.21 & 0.5813 & 19.53 & 0.4433 & 19.64 & 0.4587 & 19.56 & 0.4844\\
    EvrNet & 78.7K & 31.16 & 0.8597 & 31.22 & 0.8634 & 31.25 & 0.8630 & 29.03 & 0.8321 & 28.87 & 0.8269 & 29.17 & 0.8363\\ \midrule
    ViWSNet & 57.8M & 16.47 & 0.5243 & 16.48 & 0.5224 & 16.47 & 0.5242 & 13.83 & 0.3585 & 13.81 & 0.3598 & 13.81 &  0.3574\\
    LaverNet (Ours) & 362.7K & \textbf{32.58} & \textbf{0.8962} & \textbf{32.60} & \textbf{0.8967} & \textbf{32.54} & \textbf{0.8981} & 30.28 & \textbf{0.8859} & \textbf{30.08} & \textbf{0.8819} & 30.56 & \textbf{0.8909}\\
    \bottomrule 
    \end{tabular}
    }
\end{center}
\end{table*}

\begin{figure*}[hbtp]
    \centering
    \includegraphics[width=\linewidth]{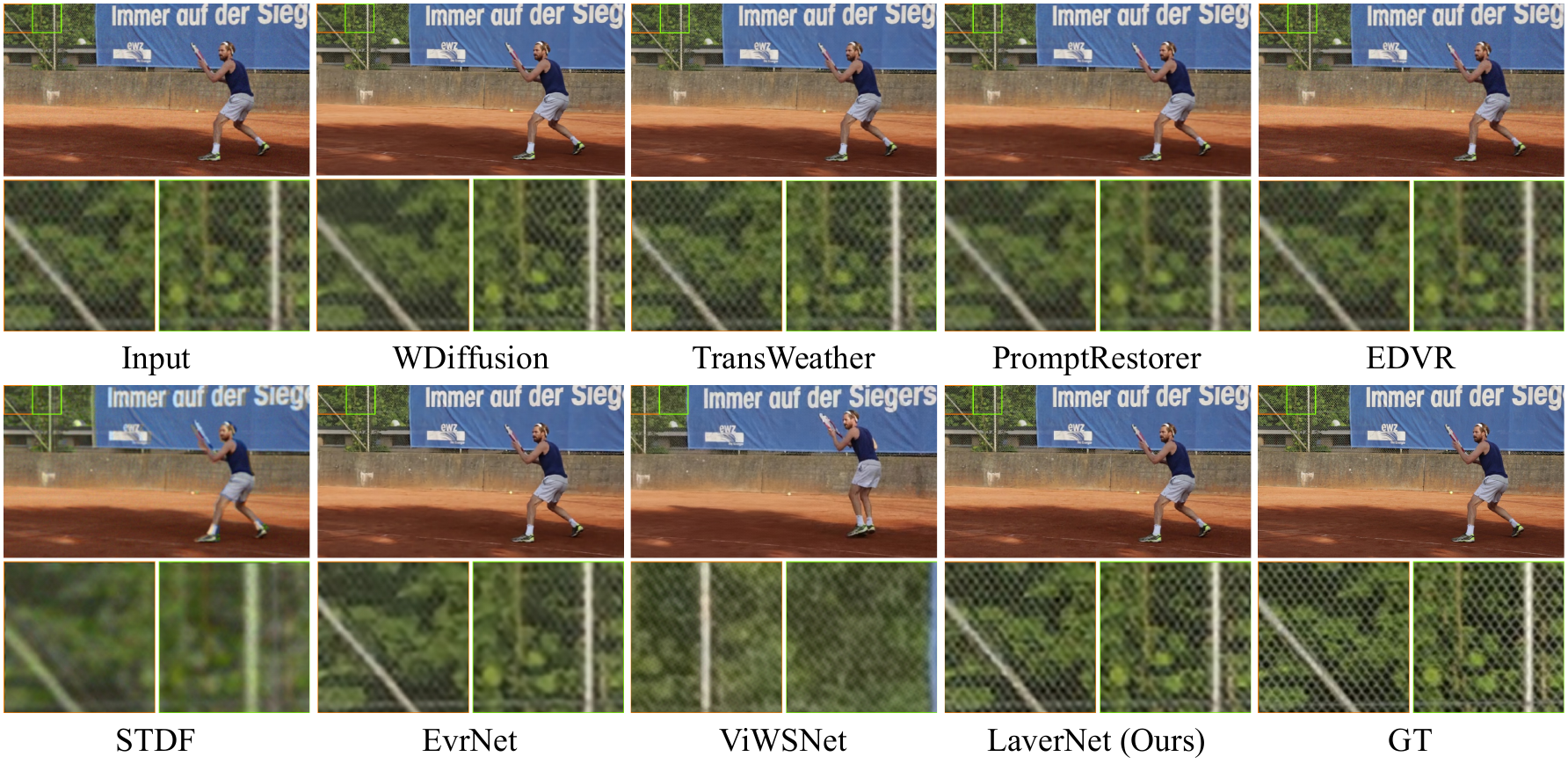}
    
    \caption{Qualitative results on the ``tennis-vest'' video from DAVIS-test ($t=12$), from which one could observe that existing methods tend to produce blurry outputs with visible artifacts. In contrast, our method effectively restores the sharp details of the wire mesh, yielding results that are closer to the ground truth.}
    \vspace{-1em}
    \label{Fig:DAVIS_t12}
\end{figure*}

To demonstrate the efficiency of our LaverNet, we compare the parameters and runtime of our method and the SOTA methods in Tab.~\ref{Tab:paramCompare}. Specifically, runtime is measured on a 48-frame video from the DAVIS-test set, while FLOPs are calculated on a single frame with a resolution of 856$\times $ 480, ensuring a fair comparison between image and video restoration methods. As shown in Tab.~\ref{Tab:paramCompare}, our method achieves the lowest number of parameters and the fastest runtime among both image restoration and all-in-one video restoration methods. Compared with the strong baseline PromptRestorer, LaverNet uses only about 1.5\% of its parameters and approximately 2.8\% of its runtime.

To evaluate LaverNet on dealing with time-varying degradations, we conduct experiments on time-varying degradations~\cite{AverNet} with different variation intervals and different degradation combinations. The variation interval refers to the interval of video degradation variations over time and degradation combinations refer to various types of degradations in the video.

\begin{table*}[htbp]
\caption{Quantitative results compared with state-of-the-art methods under three degradation combinations, namely noise\&blur, noise\&compression, and blur\&compression. The best outcomes are highlighted in \textbf{bold}.}
\label{Tab:quanti-2}
\begin{center}
    \resizebox{\textwidth}{!}{
    \begin{tabular}{l|r|cc|cc|cc|cc|cc|cc}
    \toprule
        \multicolumn{1}{c|}{\multirow{3}{*}{Method}} & \multicolumn{1}{c|}{\multirow{3}{*}{\#Param}} & \multicolumn{6}{c|}{DAVIS-test} & \multicolumn{6}{c}{Set8}  \\ \cmidrule{3-14} 
          & & \multicolumn{2}{c|}{Noise \& Blur} & \multicolumn{2}{c|}{Noise \& Comp.} & \multicolumn{2}{c|}{Blur \& Comp.} & \multicolumn{2}{c|}{Noise \& Blur} & \multicolumn{2}{c|}{Noise \& Comp.} & \multicolumn{2}{c}{Blur \& Comp.} \\ \cmidrule{3-14} 
          & & PSNR & SSIM & PSNR & SSIM & PSNR & SSIM & PSNR & SSIM & PSNR & SSIM & PSNR & SSIM \\
    \midrule
    WDiffusion & 80.0M & 32.70 & 0.8990 & 33.52 & 0.9124 & 33.76 & 0.9142 & \textbf{31.64} & 0.8943 & 30.88 & 0.8968 & 31.22 & 0.8978  \\
    TransWeather & 38.1M & 31.74 & 0.8863 & 32.53 & 0.9062 & 32.18 & 0.9017 & 29.74 & 0.8714 & 29.93 & 0.8886 & 29.61 & 0.8830  \\
	LoRA-IR & 68.9M & 25.79 & 0.6231 & 26.85 & 0.6571 & 29.27 & 0.7579 & 23.92 & 0.6056 & 23.68 & 0.6037 & 27.08 & 0.7163 \\
	PromptRestorer & 24.5M & 32.31 & 0.8893 & 33.54 & 0.9127 & 34.12 & 0.9101 & 31.03 & 0.8683 & 31.10 & 0.8994 & 30.69 & 0.8760  \\
    \midrule
    EDVR & 23.6M & 28.00 & 0.6809 & 29.58 & 0.7036 & 34.17 & 0.9082 & 27.82 & 0.7268 & 27.23 & 0.7245 & \textbf{32.15} & 0.8845  \\
    RFDA & 1.3M & 20.83 & 0.6358 & 20.76 & 0.6375 & 20.80 & 0.6393 & 19.40 & 0.5288 & 19.28 & 0.5253 & 19.34 & 0.5319\\
    STDF & 365.0K & 20.88 & 0.4947 & 20.67 & 0.4927 & 21.96 & 0.7124 & 19.49 & 0.4270 & 19.57 & 0.4431 & 20.22 & 0.5895 \\
    EvrNet & 78.7K & 31.75 & 0.8771 & 32.43 & 0.8884 & 32.08 & 0.8830 & 29.41 & 0.8382 & 29.82 & 0.8531 & 29.13 & 0.8373\\ \midrule
    ViWSNet & 57.8M & 16.48 & 0.5237 & 16.45 & 0.5226 & 16.49 & 0.5233 & 13.81 & 0.3562 & 13.73 & 0.3521 & 13.76 & 0.3563 \\
    LaverNet (Ours) & 362.7K & \textbf{33.43} & \textbf{0.9125} & \textbf{34.13} & \textbf{0.9243} & \textbf{34.23} & \textbf{0.9264} & 31.08 & \textbf{0.8957} & \textbf{31.17} & \textbf{0.9048} & 30.74 & \textbf{0.9002}\\
    \bottomrule
    \end{tabular}
    }
\end{center}
\end{table*}

\begin{figure*}
    \centering
    \includegraphics[width=\linewidth]{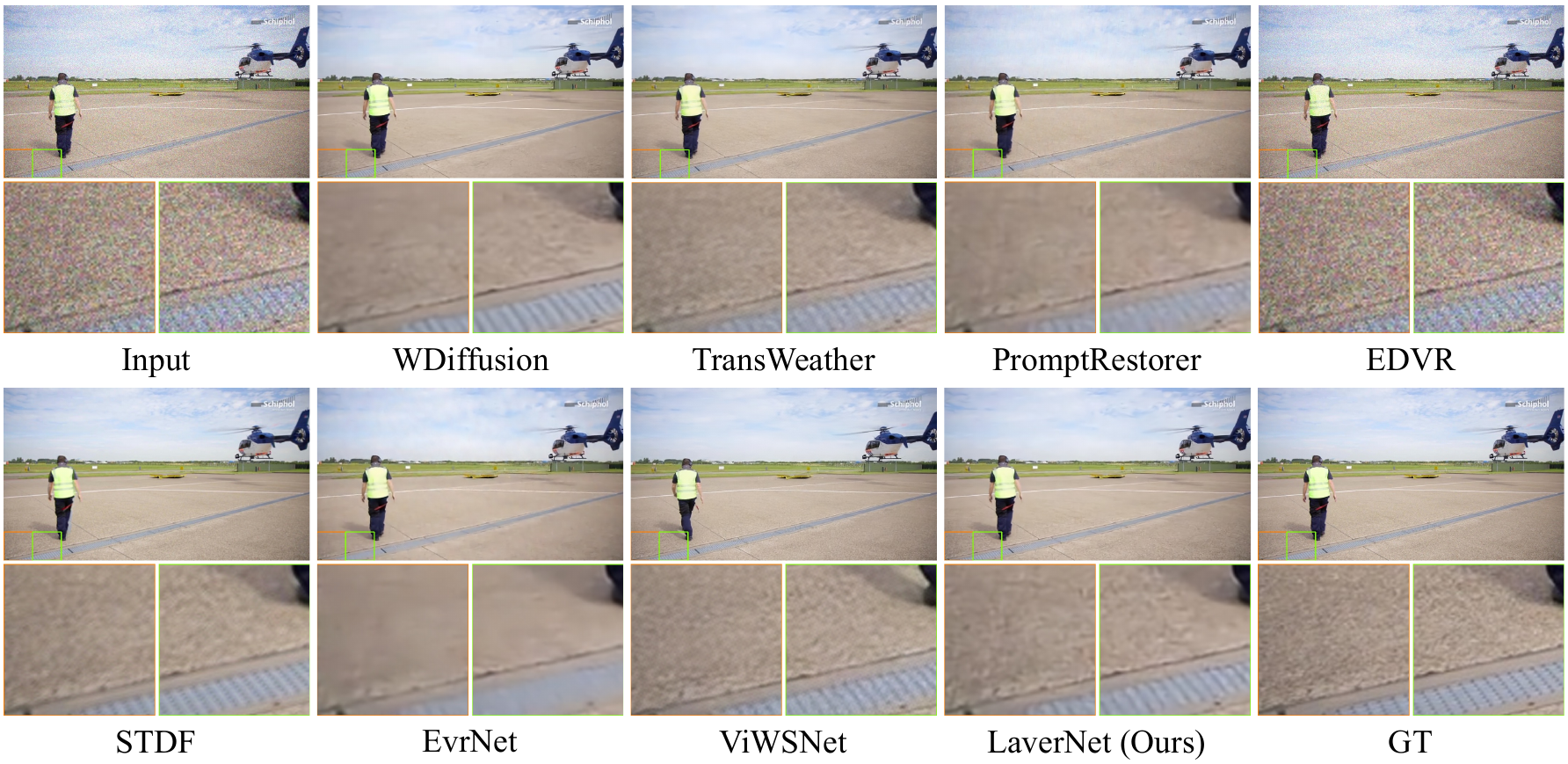}
    \caption{Qualitative results on the ``helicopter'' video from DAVIS-test in the noise\&compression degradation combination, from which one could observe that existing produce blurry or noisy results, while our method restores clearer outlines that are closer to GT.}
    \label{Fig:DAVIS_NoiseComp}
\end{figure*}

\textbf{Evaluation on Different Variation Intervals.} First, we evaluate our LaverNet and the baseline methods on six test sets with different degradation change intervals. Specifically, the test sets are synthesized on DAVIS-test and Set8 using the video synthesis approach from~\cite{AverNet}, with change intervals of $t=6,12,24$.

As shown in Tab.~\ref{Tab:quanti-1}, our method achieves comparable or even superior performance compared with the methods with significantly larger parameters. Specifically, on the DAVIS-test, it outperforms PromptRestorer by up to 1.2dB in PSNR and 0.0272 in SSIM, despite using only about 1.5\% of its parameters. Moreover, as shown in Tab.~\ref{Tab:paramCompare}, our method provides the best trade-off between performance and efficiency, requiring less than 3\% runtime of PromptRestorer and having the second lowest FLOPs among all compared methods. Compared with existing video restoration methods, our approach achieves significantly better results, demonstrating the effectiveness of selective propagation. Without the interference from degradations during temporal propagation, our methods attains outstanding performance while maintaining high efficiency. For example, LaverNet outperforms EvrNet by at least 1.68dB/0.0354 and 1.35dB/0.0517 in PSNR/SSIM on DAVIS-test and Set8, respectively.

Fig.~\ref{Fig:DAVIS_t12} presents the qualitative results on DAVIS-test dataset with interval $t=12$. It can be observed that existing image restoration methods fail to recover fine structures, leading to blurry and distorted outputs. Furthermore, the results from lightweight video restoration methods also suffer from severe blur and artifacts. In contrast, our method accurately reconstructs the fin structural details of the wire mesh, yielding sharper and more faithful results.

\textbf{Evaluation on Different Degradation Combinations.} To further evaluate the effectiveness under different degradation combinations, we test all models on the test sets synthesized in~\cite{AverNet}, which include three degradation combinations, namely noise and blur, noise and compression, and blur and compression.

As shown in Tab.~\ref{Tab:paramCompare} and~\ref{Tab:quanti-2}, our method demonstrates comparable or even superior performance while achieving significantly high efficiency. On DAVIS-test with noise and compression degradations, it outperforms EDVR by 4.55dB in PSNR and 0.2207 in SSIM, despite using only about 1.5\% of its parameters and less than one fifth of its runtime. Compared with the lightweight video restoration method EvrNet, our method achieves significantly better results, exceeding it by at least 1.68dB and 1.35dB on DAVIS-test and Set8, respectively. These results demonstrate the effectiveness of our lightweight designs in achieving efficient all-in-one video restoration.

The visualization in Fig.~\ref{Fig:DAVIS_NoiseComp} presents the qualitative results on the DAVIS-test under the combined degradation of noise and compression. As shown, other methods produce noticeably blurry results, particularly the lightweight video restoration methods STDF and EvrNet. In contrast, our method reconstructs clear outlines and yields results that are closer to the ground truth.

\subsection{Analysis Experiments}

In this section, we evaluate our model at different scales to investigate the impact of model size on restoration performance. In addition, we perform ablation studies on the key modules, namely SPM and LEM, to quantify their contributions to the overall performance.

To investigate the effect of model size on restoration performance, we construct two additional variants by adjusting the channel numbers. The tiny version, LaverNet-T, contains only 96.4K parameters, while the large version, LaverNet-L has 799.1K parameters. As shown in Tab.~\ref{Tab:modelScale}, the performance of our method consistently improves as the model scale increases. Specifically, LaverNet-L achieves the best performance with a PSNR of 32.92dB and an SSIM of 0.9027, surpassing the base model by 0.32dB and 0.0060 in PSNR and SSIM, respectively. Despite its compact size, the tiny version LaverNet-T still achieves competitive results among video restoration methods, reaching a PSNR of 31.88dB and an SSIM of 0.8831, while maintaining only 96.4K parameters.

\begin{table}[htbp]
\caption{Comparison of our models with different scales. The performance consistently improves as the model scale increases. With a parameter count comparable to EvrNet, the tiny version LaverNet-T achieves 0.69dB higher PSNR than EvrNet.}
\centering
\begin{tabular}{l|r|r|c|c}
\toprule
     & \multirow{2}{*}{\#Param} & \multirow{2}{*}{FLOPs} & \multicolumn{2}{|c}{DAVIS-test(t=12)}  \\
     \cmidrule{4-5}
     & & & PSNR & SSIM \\ \midrule
     LaverNet-T & 96.4K & 9.72G & 31.88 & 0.8831\\
     LaverNet & 362.7K & 36.90G & 32.60 & 0.8967\\
     LaverNet-L & 799.1K & 81.55G & 32.92 & 0.9027 \\
     \bottomrule
\end{tabular}
\label{Tab:modelScale}
\end{table}

\begin{table}[htbp]
\caption{Ablation studies. Each module contributes to improvements in PSNR and SSIM, validating their effectiveness.}
\vspace{-1em}
\label{Tab:ablation}
\begin{center}
\begin{tabular}{c|c|c|r|c|c}
\toprule
  & \multirow{2}{*}{SPM} & \multirow{2}{*}{LEM} & \multirow{2}{*}{\#Param} & \multicolumn{2}{c}{DAVIS-test(t=12)}  \\ \cmidrule{5-6}
  &  & & & PSNR & SSIM  \\ \midrule
(A) &            & \checkmark & 342.1K & 32.32 & 0.8896  \\
(B) & \checkmark &            & 329.5K & 32.23 & 0.8911 \\
(C) & \checkmark & \checkmark & 362.7K & 32.60 & 0.8967  \\
\bottomrule
\end{tabular}
\end{center}
\vspace{-1em}
\end{table}

To validate the effectiveness of each key module, we conduct ablation studies on SPM and LEM. Specifically, we remove the SPM from the network to examine its impact. For LEM, we replace its attention block with a dense block to examine its impact on adaptive degradation handling.

As shown in Tab.~\ref{Tab:ablation}, removing the SPM leads to a PSNR decrease of 0.28dB and an SSIM decrease of 0.0071, demonstrating that SPM plays a crucial role in temporal propagation and effectively mitigates the interference of degradations. It is worth noting that the SPM introduces only 20.6K additional parameters while contributing substantially to the overall performance. Similarly, replacing the attention block in LEM results in a PSNR drop of 0.37dB and an SSIM drop of 0.0056, indicating that the degradation-guided attention is essential for effectively handling diverse degradations.

\section{Conclusion}

In this paper, we introduced LaverNet, a lightweight all-in-one video restoration network tailored for realistic scenarios characterized by time-varying degradations. LaverNet is built upon a novel propagation mechanism that dynamically selects and transmits degradation-agnostic features across frames, enabling robust temporal modeling under complex and time-varying degradations. Together with a carefully designed lightweight enhancement module, LaverNet delivers strong restoration performance while using only about 362K parameters, providing an effective and efficient solution for all-in-one video restoration.

{\small
\bibliographystyle{ieeenat_fullname}
\bibliography{reference}
}

% WARNING: do not forget to delete the supplementary pages from your submission 
% \input{sec/X_suppl}

\end{document}